\documentclass[11pt]{article}

\usepackage{acl}
\newcommand{\ours}{Defeasible Video Entailment }

\usepackage{sidecap}

\usepackage{times}
\usepackage{latexsym}
\usepackage{booktabs}
\usepackage{booktabs}  
\usepackage{colortbl}  
\usepackage{xcolor}    
\usepackage{graphicx}  
\usepackage{makecell}  
\usepackage{array}     
\usepackage{multirow}  
\usepackage{makecell} 
\usepackage{pifont}
\usepackage{amsmath}

\newcommand{\cmark}{\ding{51}} 
\newcommand{\xmark}{\ding{55}}
\newcommand{\eat}[1]{}

\usepackage[T1]{fontenc}

\usepackage[utf8]{inputenc}

\usepackage{microtype}

\usepackage{inconsolata}

\usepackage{graphicx}

\title{Can Video Large Multimodal Models Think Like Doubters\textemdash or Double-Down: A Study on Defeasible Video Entailment}

\author{
Yue Zhang \quad Jilei Sun \quad Yunhui Guo \quad Vibhav Gogate \\
Department of Computer Science \\
The University of Texas at Dallas, Richardson, TX, USA \\
\texttt{\{yue.zhang, jilei.sun, yunhui.guo, vibhav.gogate\}@utdallas.edu}
}

\begin{document}
\maketitle
\begin{abstract}

Video Large Multimodal Models (VLMMs) have achieved impressive progress in video understanding, yet they remain limited in adaptive reasoning\textemdash the ability to revise conclusions when new evidence emerges. In reality, conclusions are rarely set in stone; additional context can strengthen or weaken an initial inference. To capture this capability, we introduce Defeasible Video Entailment (DVidE), a new task that challenges models to think like \textit{doubters}, requiring them to dynamically update entailment judgments as evidence evolves.
In DVidE, given a \textit{video premise} and a \textit{textual hypothesis}, the models must determine whether a new update strengthens or weakens the hypothesis (classification task) or generate a coherent update that modifies the entailment relationship (generation task). For solving the classification task, we propose the \textit{Chain of Counterfactual Thought} (CoCT) framework, utilizing counterfactual reasoning, ASR-enhanced video content, and rationale refinement to reduce inference bias. For the generation task, we develop a framework that combines ASR output with a Large Language Model (LLM) to produce coherent, contextually relevant updates aligned with the intended strengthener or weakener goals. 
We also introduce a benchmark dataset with human annotations and an LLM-based evaluation metric for assessing generative performance.
Experimental results show substantial improvements across multiple VLMM baselines, demonstrating that CoCT significantly enhances both classification and generation performance and strengthens the dynamic reasoning capabilities of video-language models.
\end{abstract}

\section{Introduction}
\label{sec:intro}

Video understanding has long been a central challenge in artificial intelligence, encompassing tasks such as action recognition~\cite{DBLP:conf/iccvw/MaterzynskaBBM19, DBLP:conf/cvpr/GhadiyaramTM19}, video captioning~\cite{ DBLP:conf/iccv/WangWCLWW19, DBLP:journals/corr/abs-2312-10300}, and video retrieval~\cite{DBLP:conf/iccv/HendricksWSSDR17, DBLP:conf/iccv/GaoSYN17}. 
Recent progress in Video Large Multimodal Models (VLMMs) \cite{DBLP:conf/acl/0001RKK24, DBLP:journals/corr/abs-2406-09418} has led to substantial advances in these tasks, making them primary benchmarks for assessing multi-modal visual understanding.
Yet, despite their strong ability to capture perceptual information such as objects, activities, and temporal cues \cite{DBLP:conf/eccv/LiWJ24, DBLP:journals/corr/abs-2312-14238}, they often struggle with higher-level abstract reasoning needed to fully interpret complex or ambiguous scenes. Unlike conventional video understanding tasks, video entailment \cite{DBLP:conf/cvpr/LiuCCGYYL20} evaluates whether a video premise supports or contradicts a textual hypothesis, assuming a fixed and deterministic relation between them. For example, in a sports video, if the hypothesis states, ``The player scored a goal'', the model must decide whether the visual evidence confirms or refutes that statement.

\begin{figure}[t]
    \centering
\includegraphics[width=\linewidth]{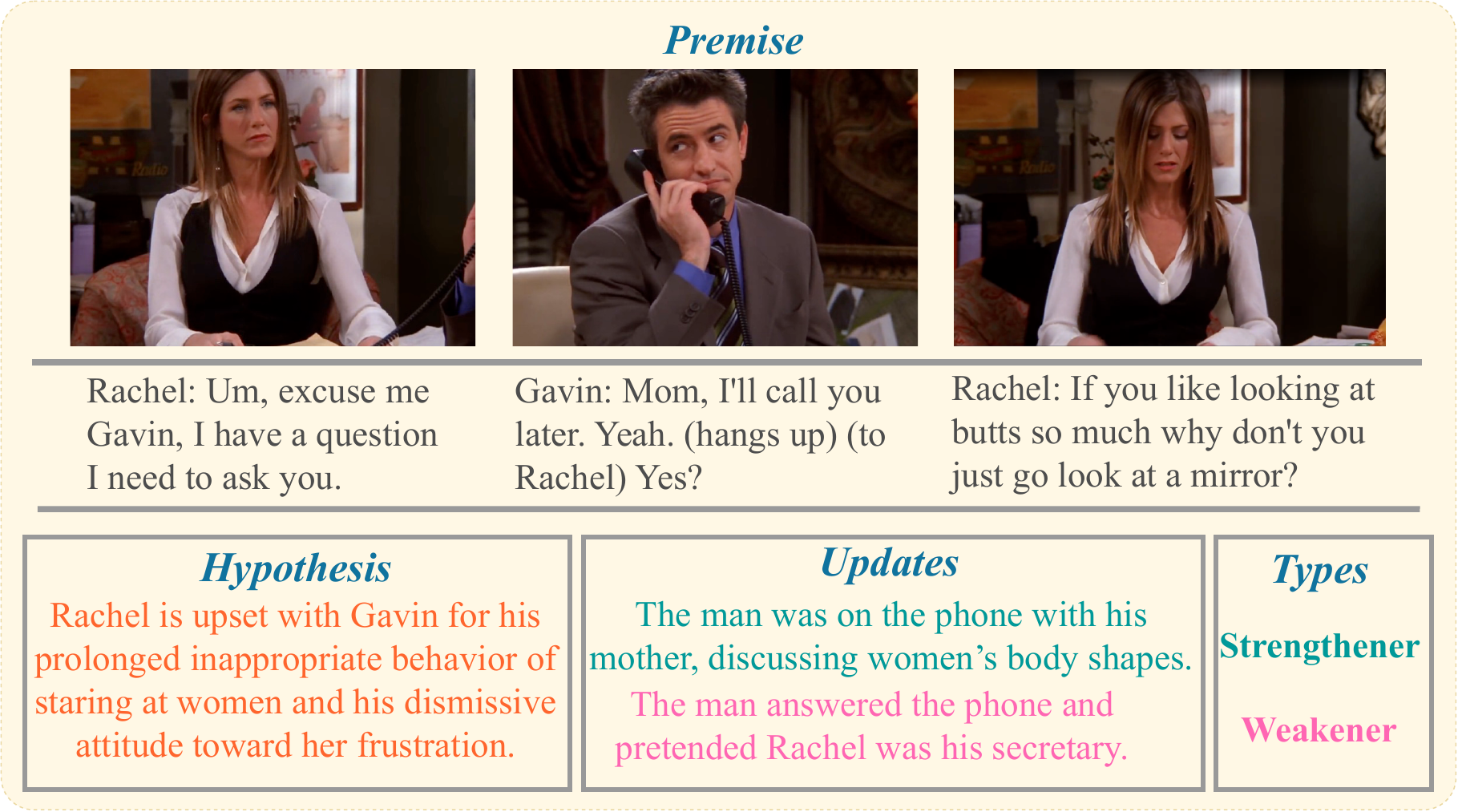}
    \caption{A defeasibility example in video entailment.}
    \vspace{-5mm}
    \label{fig:intro_case}
\end{figure}

Real-world reasoning, however, is typically defeasible, meaning conclusions can be strengthened or weakened by new information. A video clip may initially support one interpretation, but additional cues, such as dialogue obtained via Automatic Speech Recognition (ASR) or a subtle shift in visual context, may alter that conclusion. 
 As illustrated in Figure \ref{fig:intro_case}, initially, Rachel’s frustration seems to stem from Gavin’s inappropriate behavior of staring at women. However, if new information reveals that he was actually on the phone with his mother discussing women’s body shapes, this strengthens the inference. Conversely, if he was just pretending Rachel was his secretary, this could suggest that her frustration stems from being misrepresented rather than from his behavior of staring at women.

To capture this dynamic reasoning, we introduce Defeasible Video Entailment (DVidE), an extension of video entailment. Specifically, DVidE requires not only determining whether the initial relationship between a video and a hypothesis is entailment or contradiction, but also continuously revising that inference, either strengthening or weakening it, as new updates arrive. We explore this problem through two interrelated sub-tasks: (1) Classification, where the model predicts whether a given textual update strengthens or weakens the plausibility of a hypothesis given a video premise; and (2) Generation, where the model produces a contextually grounded update that meaningfully alters the entailment relation in the desired direction (strengthener or weakener).



Our initial experiments revealed that existing VLMMs perform poorly on DVidE due to three key limitations:
(1) \emph{Inability to Fully Capture Audio Information}: Current VLMMs frequently struggle with audio-related tasks, failing to accurately comprehend and interpret audio semantics in the video.
(2) \emph{Weak Visual Reasoning Ability}: Biases in VLMM training can lead to limitations in visual reasoning, resulting in generated reasoning text that lacks diversity, particularly in out-of-distribution scenarios \cite{DBLP:journals/ijcv/WangYWHCYXXZ24}.
(3) \emph{Poor Language Reasoning Ability}: Vision-based fine-tuning can sometimes degrade an LLM’s original generative and reasoning capabilities \cite{DBLP:conf/eccv/YanBCZHL24, DBLP:journals/corr/abs-2404-10710}. 

To overcome these limitations, we propose the Chain of Counterfactual Thought (CoCT) framework, which explores diverse logical relationships among the video premise, hypothesis, and update to mitigate inference bias in classification.
CoCT consists of three key components. First, Counterfactual Thought-induced Rationale Generation explores diverse reasoning processes through counterfactual reasoning, enabling VLMMs to consider multiple perspectives and improve inference diversity. Second, ASR-enhanced Video Content Integration leverages an external ASR tool to supplement VLMMs’ perception of audio information, ensuring a more comprehensive understanding of video content. Third, Rationale Refinement and Selection utilizes the strong language reasoning capabilities of LLMs to refine the generated rationales, systematically removing inference errors introduced by VLMMs, and thereby improving final classification accuracy. Additionally, we incorporate ASR-enhanced Video Content Integration to refine video descriptions before prompting an LLM to refine the update based on the hypothesis, goal, and generated update, enhancing VLMMs’ generative capabilities for our generation task. 

To evaluate VLMMs, we employ annotators to construct the \ours benchmark. In addition to traditional generation metrics, we introduce a novel Large Language Model (LLM)-based evaluation metric specifically designed for our generation task. Experimental results demonstrate that both our classification method and generation method achieve significant improvements across multiple metrics.  On the classification task,  our approach achieves an accuracy improvement of 29.7\% over the Video-ChatGPT model \cite{DBLP:conf/acl/0001RKK24}; On the generation task, our method increases the proportion of correctly generated updates from 7.50\% to 60,63\% on VideoLLaMA \cite{DBLP:journals/corr/abs-2406-07476}.

Our contributions can be summarized as:
1) We introduce a benchmark for defeasible video entailment, augmenting existing video entailment datasets with strengthener/weakener annotations and extending video-language reasoning.
2) We develop a novel Chain of Counterfactual Thought classification framework that leverages counterfactual reasoning to explore the diverse entailment semantic relations.
3) We propose a generation framework that integrates the external ASR tool and LLM to produce coherent and contextually relevant updates while ensuring they align with the goal (Weakener/Strengthener).
4) We propose an LLM-based evaluation method for our generation task. Our experimental results demonstrate that our proposed methods significantly improve both classification and generation performance on DVidE.


\section{Related Work}
\label{sec:related_work}

\textbf{Defeasible Inference Tasks.}  
Defeasible inference extends traditional entailment~\cite{DBLP:conf/naacl/BosM05, DBLP:conf/mlcw/DaganGM05, DBLP:conf/iwcs/MacCartneyM09, DBLP:conf/emnlp/BowmanAPM15} by considering how new information can make the previous hypothesis more or less likely. Although standard Natural Language Inference (NLI) datasets, like SNLI~\cite{DBLP:conf/emnlp/BowmanAPM15} and MNLI~\cite{DBLP:conf/naacl/WilliamsNB18}, focus on fixed premise-hypothesis relationships, recent work has explored defeasibility, where a piece of new information can strengthen or weaken the hypothesis. For example, the $\delta$-NLI dataset \cite{DBLP:conf/emnlp/RudingerSHBFBSC20} specifically studies changes in entailment by providing additional textual information that makes the hypothesis more or less plausible. Similarly, $\delta$-CAUSAL~\cite{DBLP:journals/corr/abs-2401-03183} extends defeasibility to causal reasoning, evaluating how new facts influence inference dynamics. While these datasets introduce text-based defeasible inference, they do not incorporate multimodal reasoning. SNLI-VE~\cite{DBLP:journals/corr/abs-1901-06706} extends textual entailment to the visual domain by determining whether a hypothesis is entailed or contradicted by an image. 
The DVE dataset~\cite{DBLP:conf/aaai/ZhangJG25} introduces defeasibility based on the SNLI-VE, allowing new textual information to alter image-hypothesis relationships. 
However, DVE remains limited to single-frame contexts, lacking the temporal visual and auditory information for models to judge whether the new information strengthen or weaken the hypothesis. Our work extends defeasible inference to video, providing models with richer temporal and multimodal context to improve understanding.

\textbf{Video Understanding Tasks.} 
Video understanding covers a range of tasks, such as activity recognition, event localization, and video-language reasoning. Activity recognition~\cite{DBLP:journals/corr/abs-1212-0402, DBLP:conf/iccv/KuehneJGPS11, DBLP:conf/cvpr/PiergiovanniR18a, DBLP:conf/iros/RoitbergSDSRS21} focuses on identifying human actions within videos. Event localization~\cite{DBLP:conf/cvpr/HeilbronEGN15, DBLP:journals/cviu/IdreesZJGLSS17, DBLP:conf/eccv/TianSLDX18, DBLP:conf/cvpr/GengW0CZ23} aims to detect the temporal boundaries of specific events, which is essential for video summarization and highlight detection. In addition, Video-language reasoning~\cite{DBLP:conf/emnlp/LeiYBB18, DBLP:conf/cvpr/XiaoSYC21, DBLP:conf/cvpr/XuMYR16, DBLP:conf/iccv/MiechZATLS19} involves integrating textual and visual information to answer questions, generate descriptions, or make logical inferences about video content. While these tasks contribute significantly to video understanding, they primarily focus on static interpretation—classifying actions, detecting events, or retrieving relevant textual descriptions—without modeling how new evidence alters existing inferences. In real-world settings, reasoning is inherently defeasible, requiring conclusions to adapt as new evidence emerges. However, current benchmarks rarely capture such dynamic inference, limiting their relevance for tasks that require continual reasoning updates.

\section{Problem Formulation}
Following the definition outlined in \cite{DBLP:conf/emnlp/RudingerSHBFBSC20}, we define our \ours  task as follows
\begin{quote}
    Given a video clip premise $V$, a hypothesis $ H $ is defeasible if there exists an update $ {U} $ (consistent with $ {V} $) such that a human would find $ {H} $ less likely to be true after learning $ {U} $. 
    Specifically, an update $ {U} $ is called a weakener if given a premise $ {V} $ and hypothesis $ {H} $, a human would most likely find $ {H} $ less likely to be true after learning $ {U} $; if they find $ {H} $ more likely to be true, then we call $ {U} $ a strengthener. 
    Importantly, the update must remain consistent with the video premise: it may modify belief in the hypothesis but cannot contradict the observed evidence.
\end{quote}

\noindent\textbf{Classification Task.}
The goal is to find a classification model $\mathcal{M}_c$ which predicts the update type based on the premise $V$, hypothesis $H$, and update $U$, formulated as  $\hat{L} = \mathcal{M}_c (V, H, U)$, 
where $\hat{L} \in \{w, s, h\}$ denotes the predicted update type. 
Specifically, $\hat{L} = s$ (strengthener) is assigned if $U$ makes the hypothesis $H$ more likely to be true given the video clip $V$, and $\hat{L} = w$ (weakener) is assigned if $U$ makes the hypothesis $H$ less likely to be true given $V$.

\noindent\textbf{Generation Task.} 
\label{sec:gen}
The objective is to train a model $\mathcal{M}_g$ that generates a textual update $\hat{U}$ based on the input video clip $V$, hypothesis $H$, and goal $G \in \{w,s\}$ (i.e.\ weakener or strengthener), formulated as $\hat{U} = \mathcal{M}_g (V,H,G)$. 

\section{Defeasible Video Entailment Benchmark}

\subsection{Data Source} 

To build our \ours benchmark, we start with VIOLIN \cite{DBLP:conf/cvpr/LiuCCGYYL20}, a large-scale video-and-language inference dataset offering labeled video-hypothesis pairs for entailment classification. We then augment these pairs with textual updates that either strengthen or weaken the hypothesis, thus introducing the concept of defeasibility into video-based reasoning.

\begin{table}[t]
\centering
\small
\vspace{-3mm}
\begin{tabular}{ll}
\toprule
\textbf{Statistics} & \textbf{Test set} \\ 
\midrule
Total samples              & 986  \\ 
Update type distribution   &       \\ 
\quad Weakener             & 493  \\ 
\quad Strengthener         & 493  \\ 
Average hypothesis length  & 17.99 words \\ 
Average update length      & 12.28 words \\ 
Average video duration     & 16.45 sec   \\ 
Unique videos              & 986  \\ 
\bottomrule
\end{tabular}
\vspace{-3mm}
\caption{Statistics of the DVidE dataset.}
\label{tab:data_stats}
\end{table}

Specifically, VIOLIN consists of 95,322 \textbf{(video premise, text hypothesis)} pairs from 15,887 video clips, spanning over 582 hours of video. The videos are sourced from two domains: 1) popular TV shows and 2) movie clips from YouTube channels. Therefore, the topics of the video source are very comprehensive. 
For each video clip, the annotators wrote multiple hypotheses that can be categorized as real statements or fake statements.
A hypothesis is considered real if it is supported by the video or fake if it contradicts the video’s content.
To incorporate defeasibility into the video domain, we extend VIOLIN by introducing annotations and adjusting video clips to construct (video premise, text update, text hypothesis) triplets. The detailed process is given in the next section.

\subsection{Data Annotation}
To construct our benchmark, we first randomly select 200 videos from VIOLIN and gather their corresponding annotated hypotheses for manual annotation. Because many of VIOLIN's video clips include clear evidence either supporting or contradicting the hypotheses, they are inherently non-neutral and thus unsuitable for defeasible reasoning. To address this, we focus on creating neutral (video premise, hypothesis) pairs in which the video premise neither strongly supports nor outright contradicts the hypothesis. 


To create neutral (video premise, hypothesis) pairs, we remove the sub-clip evidence in the video premise that directly supports or contradicts the hypothesis, thereby removing explicit evidence.
Specifically, if the sub-clip evidence appears at the beginning or end of the video, the remaining portion is retained as the premise. If the evidence is located in the middle, we select the longer of the two remaining segments to maintain coherence. The hypotheses from VIOLIN are used as our hypotheses, while the premises are derived from the edited video segments with the evidence removed.
The edited video clips ensure that the premise neither inherently supports nor contradicts the statement, making it neutral.

\begin{table}[t]
\centering
\resizebox{\linewidth}{!}{%
\begin{tabular}{lcccc}
\toprule
\textbf{Dataset} & \textbf{Multimodal} & \textbf{Video} & \textbf{Strengthener} & \textbf{Weakener} \\
\midrule
\rowcolor{gray!10} \multicolumn{5}{c}{Video Understanding Datasets} \\
Charades~\cite{DBLP:conf/eccv/SigurdssonVWFLG16} & \cmark & \cmark & \xmark & \xmark \\
InfiniBench~\cite{DBLP:journals/corr/abs-2406-19875} & \cmark & \cmark & \xmark & \xmark \\
Kinetics~\cite{DBLP:journals/corr/KayCSZHVVGBNSZ17} & \cmark & \cmark & \xmark & \xmark \\
\midrule
\rowcolor{gray!10} \multicolumn{5}{c}{Natural Language Inference Datasets} \\
SNLI~\cite{DBLP:conf/emnlp/BowmanAPM15}  & \xmark & \xmark & \xmark & \xmark \\
MNLI~\cite{DBLP:conf/naacl/WilliamsNB18} & \xmark & \xmark & \xmark & \xmark \\
$\delta$-NLI~\cite{DBLP:conf/emnlp/RudingerSHBFBSC20} & \xmark & \xmark & \cmark & \cmark \\
$\delta$-CAUSAL~\cite{DBLP:journals/corr/abs-2401-03183} & \xmark & \xmark & \cmark & \cmark \\
SNLI-VE~\cite{DBLP:journals/corr/abs-1901-06706} & \cmark & \xmark & \xmark & \xmark \\
DVE~\cite{DBLP:conf/aaai/ZhangJG25} & \cmark & \xmark & \cmark & \cmark \\
\midrule
DVidE & \cmark & \cmark & \cmark & \cmark \\
\bottomrule
\end{tabular}
}
\caption{Comparison of DVidE with related datasets.}
\vspace{-3mm}
\label{tab:dataset_comparison}
\end{table}

To simplify the updated annotation process, we directly ask annotators to write textual updates based on the removed sub-clip, allowing our neutral (video premise, hypothesis) pairs to become real or fake statements. This is because the textual information serves as a substitute for the removed sub-clips semantics, providing direct evidence for entailment relation inference.
For real statements, annotators are asked to manually write updates (strengtheners) based on the removed sub-clips to make the hypothesis more likely. Similarly, for false hypotheses, we construct weakening updates based on the removed sub-clips to make the hypothesis less likely to hold. This annotation strategy ensures that each update is grounded in the omitted portion of the video while remaining consistent with the original premise.

To enhance the reliability of our data, each sample in our benchmark is annotated by three independent annotators. Specifically, the first annotator generates an update based on our neutral premise-hypothesis pair and the corresponding removed sub-clip.
A second annotator reviews each update to verify that it satisfies the required criteria and provides corrections if discrepancies are found. Finally, a third annotator performs a final pass, making additional adjustments as needed to ensure overall accuracy and consistency.

\begin{figure*}[t]
    \centering
    \includegraphics[width=0.9\linewidth]{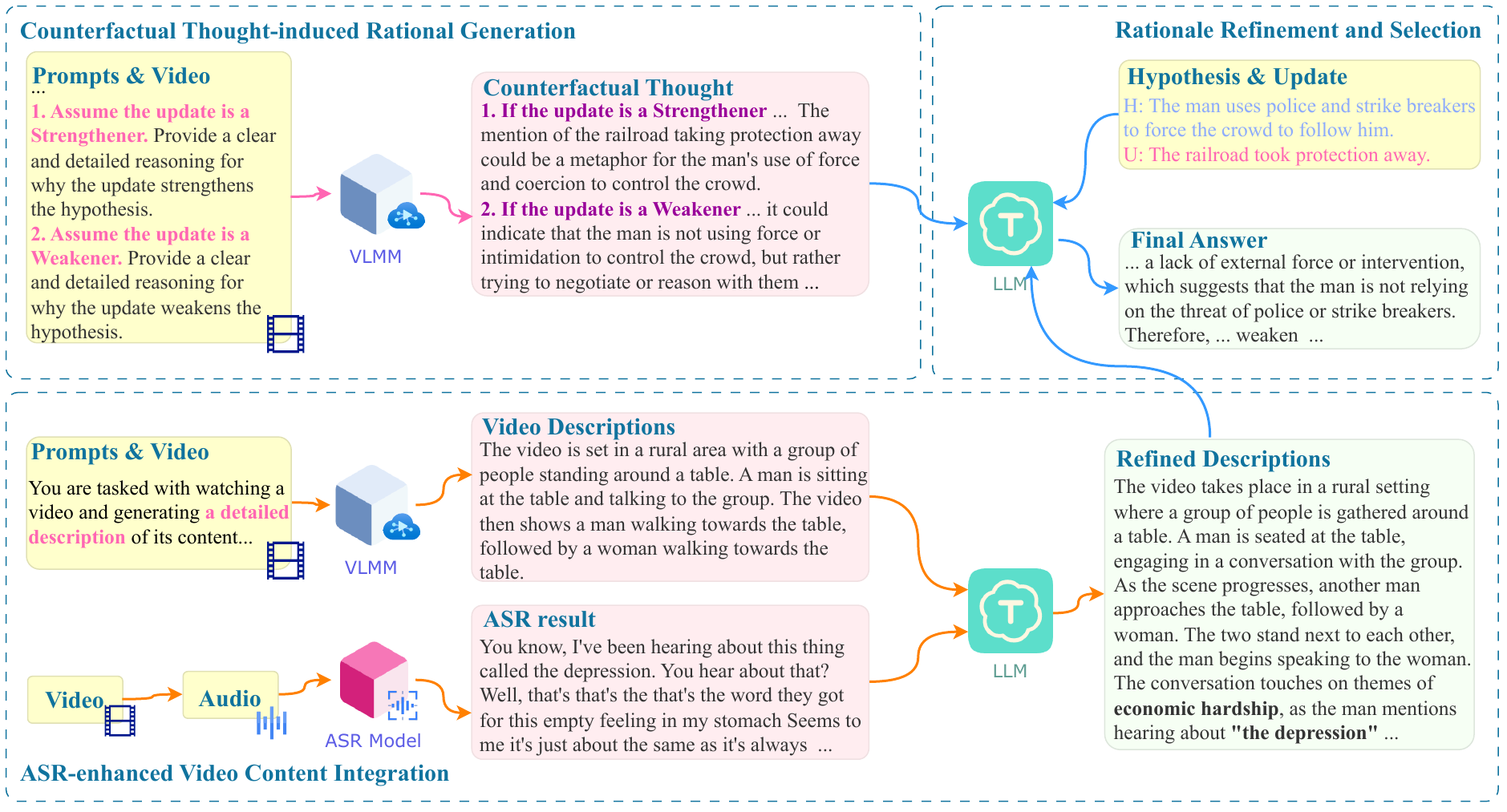}
    \caption{The architecture of Chain of Counterfactual Thought Classification Framework, including three modules: Counterfactual Thought-Induced Rational Generation, ASR-Enhanced Video Content Integration, and Rationale Refinement and Selection.}
    \label{fig:method_cls}
    \vspace{-2mm}
\end{figure*}

\subsection{Statistics and Comparison}
The statistics of our benchmark are summarized in Table \ref{tab:data_stats}. Overall, our dataset’s balanced and diverse data support comprehensive training and evaluation of models on defeasible video entailment tasks.
We compare our DVidE dataset with related datasets and categorize them into two groups: video understanding datasets, and natural language inference datasets. Table \ref{tab:data_stats} provides a detailed comparison. More explanation about the dataset comparison can be found in the Appendix \ref{app:comparison}.

\section{Improve Classification Ability of VLMMs}




In our preliminary experiments, we observed that VLMMs struggle to capture the semantic shift introduced by new updates. To address this, we integrate counterfactual reasoning to explore the diverse logical and semantic relationships among the update, premise, and hypothesis, thereby mitigating inherent inference biases. Specifically, we propose a Chain of Counterfactual Thought Classification framework that encapsulates three key modules—\textit{Counterfactual Thought-induced Rationale Generation}, \textit{ASR-enhanced Video Content Integration}, and \textit{Rationale Selection and Refinement}—as illustrated in Fig.~\ref{fig:method_cls}.


\subsection{Counterfactual Thought-Induced Rational Generation}
To capture the complexity of logical semantic relationships among the update, premise, and hypothesis, we leverage VLMMs to generate two rationales—one explaining why the update acts as a Strengthener and the other explaining why it serves as a Weakener. This approach enables counterfactual reasoning~\cite{DBLP:conf/emnlp/HeYFFAJNBW022, DBLP:conf/acl/ChenGBS023, DBLP:conf/acl/WeinzierlH24} by examining both perspectives for each case as follows,
\begin{equation}
    R_w, R_s = \operatorname{VLMM}(P(H,U), V),
\end{equation}
$P(H,U)$ is the prompt (see Appendix \ref{subsec:prompt_our_cls} for details) and $R_w$ and $R_s$ represent the rationales generated by the model, where  $R_w$  explains why the update weakens the hypothesis, and $R_s$  explains why it strengthens it. 
Our method prompts the model to analyze the update from two opposing perspectives: (1) assuming it strengthens the hypothesis and (2) assuming it weakens the hypothesis.






\subsection{ASR-Enhanced Video Content Integration}
After generating the two rationales, we must determine which one is more plausible as the final answer. However, because vision-based fine-tuning can sometimes impair an LLM’s original generative or reasoning capabilities \cite{DBLP:conf/eccv/YanBCZHL24, DBLP:journals/corr/abs-2404-10710}, we rely on a separate LLM to evaluate which rationale is stronger. Since this LLM cannot directly process visual information, its selection process may introduce biases or yield suboptimal results. To address this, we convert the video content into a textual description, enabling the LLM to incorporate the visual context when comparing the two rationales. Specifically, we input the video into VLMMs to generate the textual description, 
formulated as 
    $D' = \operatorname{VLMM}(V, P_d)$, 
where $D'$ is the generated description. $P_D$ is the prompt, as shown in Appendix \ref{subsec:prompt_our_cls}.

However, existing VLMMs often generate incomplete or inaccurate video descriptions, as they primarily rely on visual inputs and may overlook crucial auditory information~\cite{DBLP:journals/corr/abs-2401-09774}. 
This issue is especially problematic in \ours task since the spoken dialogue conveys key semantic details. 
To address this, we refine video descriptions by integrating VLMM-generated description with ASR transcripts, ensuring a more comprehensive and context-aware representation of the video. 
Specifically, we first extract spoken dialogue from the video via an ASR engine to capture verbal context, and then employ an LLM to integrate the ASR transcript with the VLMM-generated description as follows,
    \begin{align}
        &T_A = \operatorname{ASR}(A) ,\\
        &D = \operatorname{LLM}(P_A(T_A, D')), 
    \end{align}
where $D$ is the refined description and $A$ is the audio.  $T_A$ is the ASR text. $P_A(T_A, D')$ is the prompt, as shown in Appendix \ref{subsec:prompt_our_cls}.
This refinement is particularly important in scenarios where key details—such as causal relationships between events or character motivations—are conveyed primarily through dialogue rather than visual cues.

\begin{figure}[t]
    \centering
    \includegraphics[width=\linewidth]{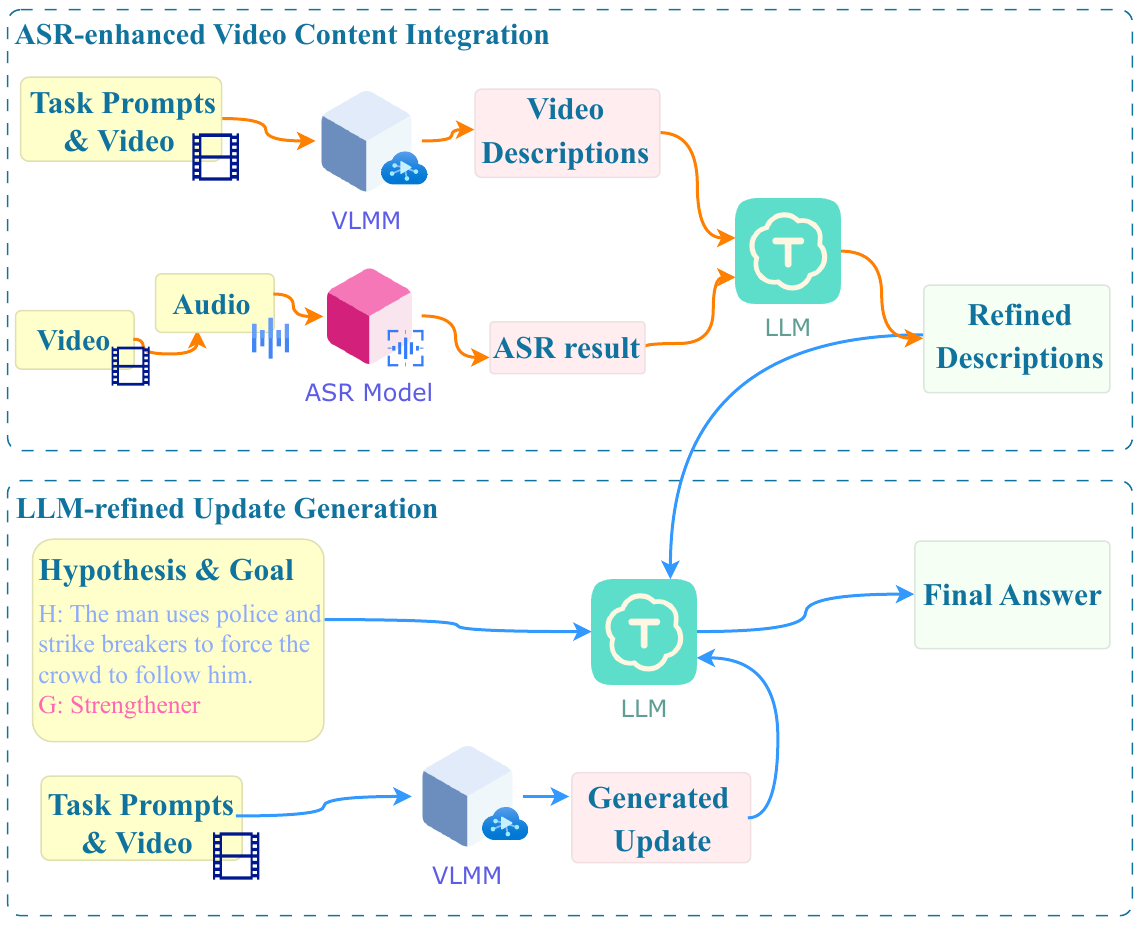}
    \caption{The architecture of LLM-Guided ASR-Integrated Generation Framework, including ASR-Enhanced Video Content Integration, and LLM-Refined Update Generation.}
    \label{fig:method_gen}
\end{figure}

\subsection{Rationale Refinement and Selection}
As we mentioned before, because vision-based fine-tuning can sometimes impair an LLM’s original generative or reasoning abilities, we rely on a separate LLM to evaluate which rationale is stronger. In addition, we occasionally observe that the rationales for both weakener and strengthener are not reasonable. Therefore, we also ask the LLM to refine the rationale by revising reasoning errors before it outputs the final answer, 
formulated as 
    $F = \operatorname{LLM}(P_F(R_s,R_w,D,H,U))$,
$P_F(R_s,R_w,D,H,U)$ is the prompt detailed in Appendix \ref{subsec:prompt_our_cls}. $F$ is the final result, including the refined rationals and the selection result.


\section{Improve Generation Ability of VLMMs}


As noted earlier, current VLMMs face two key limitations: (1) they often miss crucial audio information, and (2) they fail to fully capture the entailment semantics among the update, hypothesis, and premise. To address these issues, we introduce an LLM-Guided ASR-Integrated Generation Framework, which builds upon the modules developed in our classification framework (Section~\ref{sec:gen}). Specifically, we first refine the VLMM-generated descriptions through ASR-Enhanced Video Content Integration, producing more accurate textual representations of the video content. This step follows the same refinement procedure used in the classification task.

Similar to the classification framework, we use the LLM to refine VLMM-generated updates based on these refined descriptions, premise, hypothesis,  and goal to further refine the update as 
    $U = \operatorname{LLM}(P_U(D, H, P, G))$,
$P_U(D, H, P, G)$ is the prompt, as shown in Appendix \ref{subsec:prompt_our_gen}. $D$ is the refined description. $U$ is the generated update. By leveraging LLM, our method can revise the reasoning errors in VLMMs and capture the real entailment semantics. The overall architecture of the model is illustrated in
Figure \ref{fig:method_gen}.

\section{Experiment}
\label{benchvlmm}

\begin{table}[t]
\centering
\small
\resizebox{\linewidth}{!}{%
\begin{tabular}{lcccccc}
\toprule
\textbf{Model} & \textbf{w/Ours} & \textbf{Accuracy} $\uparrow$ & \textbf{Precision} $\uparrow$ & \textbf{Recall} $\uparrow$ & \textbf{F1} $\uparrow$ & \textbf{SRatio} $\leftrightarrow$ \\
\midrule
\multirow{2}{*}{VideoLLaMA\_V} & \xmark & 52.74 & 54.26 & 34.89 & 42.47 & 32.15 \\
 & \cmark & \textbf{76.55} & \textbf{80.47} & \textbf{70.18} & \textbf{74.97} & \textbf{43.71} \\
\midrule
\multirow{2}{*}{VideoLLaMA\_AV} & \xmark & 55.07 & 55.90 & 48.07 & 51.69 & \textbf{43.00} \\
 & \cmark & \textbf{77.48} & \textbf{83.13} & \textbf{68.97} & \textbf{75.39} & 41.48 \\
\midrule
\multirow{2}{*}{Video-ChatGPT} & \xmark & 61.05 & 75.59 & 32.66 & 45.61 & 21.60 \\
 & \cmark & \textbf{77.79} & \textbf{84.77} & \textbf{67.75} & \textbf{75.31} & \textbf{39.95} \\
\midrule
\multirow{2}{*}{VideoGPT+} & \xmark & 72.76 & \textbf{78.05} & 63.49 & 70.02 & 40.71 \\
 & \cmark & \textbf{77.16} & 75.87 & \textbf{79.72} & \textbf{77.74} & \textbf{43.65} \\
\midrule
\multirow{2}{*}{VideoLLaMA2\_V} & \xmark & 59.84 & 57.23 & \textbf{77.89} & 65.98 & 68.05 \\
 & \cmark & \textbf{77.89} & \textbf{82.51} & 70.79 & \textbf{76.20} & \textbf{43.40} \\
\midrule
\multirow{2}{*}{VideoLLaMA2\_AV} & \xmark & 52.94 & 51.54 & \textbf{98.17} & 67.60 & 95.23 \\
 & \cmark & \textbf{78.80} & \textbf{83.18} & 72.21 & \textbf{77.31} & \textbf{42.90} \\
\midrule
\multirow{2}{*}{Qwen2.5-VL} & \xmark & 71.10 & 73.61 & 71.10 & 70.30 & \textbf{36.71} \\
 & \cmark & \textbf{77.18} & \textbf{78.01} & \textbf{77.18} & \textbf{77.01} & 36.21 \\
\midrule
\multirow{2}{*}{InternVL3} & \xmark & 74.65 & 76.74 & 74.65 & 74.14 & \textbf{44.52} \\
 & \cmark & \textbf{76.67} & \textbf{77.36} & \textbf{76.67} & \textbf{76.53} & 33.57 \\
\bottomrule
\end{tabular}
}
\caption{Performance comparison of different VLMMs on the classification task. SRatio stands for the ratio of Strengthener. TV stands for Text and Vision modalities, and TVA stands for the Text, Vision, and Audio modalities. $\uparrow$ indicates that a higher value is better, while $\leftrightarrow$ indicates that values closer to 50 are preferable. 
}
\label{table:cls}
\end{table}

\subsection{Experimental Setup}
We select eight widely used VLMMs as baselines, categorized into two groups: Video-Only Models, which can process video but do not support audio comprehension, including Video-ChatGPT~\cite{DBLP:conf/acl/0001RKK24}, VideoGPT+~\cite{DBLP:journals/corr/abs-2406-09418}, VideoLLaMA\_V (Vision-Language Branch only) \cite{DBLP:conf/emnlp/ZhangLB23}, VideoLLaMA2\_V (Vision-Language Branch only) \cite{DBLP:journals/corr/abs-2406-07476}, Qwen2.5-VL~\cite{DBLP:journals/corr/abs-2502-13923} and internVL3~\cite{DBLP:journals/corr/abs-2504-10479}; and Video-Audio Models, which can process both video and audio, including VideoLLaMA\_AV (incorporating both Vision-Language and Audio-Language Branches) and VideoLLaMA2\_AV (incorporating both Vision-Language and Audio-Language Branches). All the VLMMs have 7B parameters.
The prompts remain consistent across all VLMMs, with details provided in Appendix \ref{subsec:prompt_cls_base}.
For the ASR engine, we use whisper \cite{DBLP:conf/icml/RadfordKXBMS23}.

\subsubsection{Evaluation Metric}
To evaluate the performance of the model $\mathcal{M}_c$ on the \textbf{classification task}, we use multiple typical classification metrics: F1, accuracy, recall, and precision. This evaluation framework provides a balanced and detailed view of the model's classification capabilities across multiple dimensions.


While traditional string-matching metrics such as ROUGE-L \cite{lin2004rouge} and BLEU-4 \cite{DBLP:conf/acl/PapineniRWZ02}, as well as embedding-based metrics like BERTScore \cite{DBLP:conf/iclr/ZhangKWWA20}, are commonly used to evaluate text generation, they primarily capture lexical or shallow semantic similarity. For open-domain tasks, where acceptable outputs can vary widely and it is impractical to construct exhaustive reference sets, these metrics are inadequate. 
To address this, we designed a set of carefully defined categories to analyze generated updates, focusing on their interactions with the premise and hypothesis. 
Specifically, we define eight categories as follows:
\textbf{C1 Good:} The update strengthens the hypothesis if the target update type is Strengthener, or weakens the hypothesis if the target update type is Weakener.
\textbf{C2 Neutral:} The update neither strengthens nor weakens the hypothesis.
\textbf{C3 Weakener instead of Strengthener:} The update, generated as a Strengthener, actually weakens the hypothesis.
\textbf{C4 Strengthener instead of Weakener:} The update, generated as a Weakener, actually strengthens the hypothesis.
\textbf{C5 Restating the Premise:} The update simply describes what is shown in the video, and does not contribute toward evaluating the hypothesis.
\textbf{C6 Restating the Hypothesis:} The update roughly repeats the hypothesis.
\textbf{C7 Contradicting the Premise:} The update contradicts the premise, either explicitly or implicitly.
\textbf{C8 Nonsensical or Other:} The update itself is nonsensical, incoherent, or unrelated.

\begin{table}[t]
    \centering
    \resizebox{0.5\textwidth}{!}{
    \begin{tabular}{lcc|cccccccc}
        \toprule
        \textbf{Model} & 
        \textbf{w/Ours} &
        \textbf{C1} $\uparrow$& 
        \textbf{C2} $\downarrow$& 
        \textbf{C3} $\downarrow$& 
        \textbf{C4} $\downarrow$& 
        \textbf{C5} $\downarrow$& 
        \textbf{C6} $\downarrow$& 
        \textbf{C7} $\downarrow$& 
        \textbf{C8} $\downarrow$\\
        \midrule
        \multirow{2}{*}{VideoChatGPT} &\xmark & 39.29 & 11.12 & 17.08 & \textbf{18.65} & \textbf{0.00} & 7.91 & 0.36 & 5.60 \\
        &\cmark & \textbf{64.86} & \textbf{0.83} & \textbf{5.89} & 23.72 & 0.00 & \textbf{4.16} & \textbf{0.00} & \textbf{1.05} \\
        \midrule
        \multirow{2}{*}{VideoGPT+} &\xmark & 35.74 & 4.38 & \textbf{5.99} & \textbf{23.68} & 0.06 & 28.91 & 0.06 & \textbf{1.18} \\
        &\cmark & \textbf{52.78} & \textbf{1.46} & 7.23 & 29.24 & \textbf{0.00} & 7.87 & \textbf{0.00} & 1.42 \\
        \midrule
        \multirow{2}{*}{VideoLLaMA\_V} & \xmark & 7.50 & 29.31 & 7.94 & \textbf{7.23} & 1.30 & 12.31 & \textbf{0.00} & 34.41 \\
        &\cmark & \textbf{60.63} & \textbf{1.03} & \textbf{5.98} & 27.20 & \textbf{0.00} & \textbf{4.26} & 0.03 & \textbf{0.86} \\
        \midrule
        \multirow{2}{*}{VideoLLaMA\_AV} &\xmark & 13.96 & 23.06 & 9.58 & \textbf{9.19} & 0.92 & 17.36 & \textbf{0.00} & 25.93 \\
        &\cmark & \textbf{63.20} & \textbf{1.40} & \textbf{6.02} & 23.79 & \textbf{0.04} & \textbf{4.77} & 0.04 & \textbf{0.75} \\
        \midrule
        \multirow{2}{*}{VideoLLaMA2\_V} &\xmark & 48.68 & 9.88 & 15.40 & \textbf{14.77} & \textbf{0.00} & 8.36 & \textbf{0.00} & 2.91 \\
        &\cmark & \textbf{56.99} & \textbf{1.41} & \textbf{4.68} & 29.33 & \textbf{0.00} & \textbf{6.67} & 0.03 & \textbf{0.90} \\
        \midrule
        \multirow{2}{*}{VideoLLaMA2\_AV} &\xmark & 47.04 & 11.53 & 14.67 & \textbf{14.41} & \textbf{0.00} & 9.39 & \textbf{0.07} & 2.89 \\
        &\cmark & \textbf{58.76} & \textbf{1.08} & \textbf{4.21} & 28.71 & \textbf{0.00} & \textbf{6.33} & \textbf{0.07} & \textbf{0.91} \\
        \midrule
        \multirow{2}{*}{Qwen2.5-VL} & \xmark & 62.17 & 30.02 & 0.66 & 0.10 & 0.00 & 0.41 & 0.56 & 6.08 \\
        & \cmark & \textbf{83.27} & \textbf{8.32} & \textbf{0.15} & \textbf{0.00} & \textbf{0.00} & 0.51 & 0.56 & \textbf{7.20} \\
        \midrule
        \multirow{2}{*}{InternVL3} & \xmark & 67.65 & 23.12 & 0.86 & 0.05 & 0.00 & 0.46 & 0.86 & 7.00 \\
        & \cmark & \textbf{80.27} & \textbf{11.51} & \textbf{0.20} & \textbf{0.00} & \textbf{0.00} & 0.46 & 0.61 & \textbf{6.95} \\
        \bottomrule
    \end{tabular}
    }
    \caption{Evaluation of VLMMs on the generation task. Each number in C1--C8 (\%) indicates the proportion of model-generated updates that fall into the category.}
    \label{table:gen_combined}
\end{table}

These categories give a detailed assessment of the overall update quality, the specific error patterns, and their alignment with the intended update type. Since human evaluation is costly, we leverage GPT-4o to categorize the generated output into one of the eight predefined categories. Specifically, we provide GPT-4o with the video description, the hypothesis, and the generated update, allowing it to determine the appropriate category. The prompt is in Appendix \ref{subsec:prompt_eval}.

\subsection{Main Results}
\textbf{Classification Task.} We show the {classification performance} comparison across baselines in Table~\ref{table:cls}. We have several observations.  1) Incorporating audio information enhances model performance, as models with audio input consistently achieve higher F1 scores. For example, the F1 of VideoLLaMA improves from 42.47\% to 51.69\% with the addition of audio, while VideoLLaMA2 sees a smaller yet notable increase from 65.98\% to 67.60\%. 
2) Newer generations of models demonstrate improved performance, with more recent models such as VideoGPT+ and VideoLLaMA2 outperforming their predecessors, VideoGPT and VideoLLaMA\_V. Specifically, VideoGPT+ achieves an F1 of 70.02\%, significantly surpassing Video-ChatGPT (45.61\%), while VideoLLaMA2\_V attains 65.98\%, outperforming the earlier VideoLLaMA\_V model (42.47\%). The Latest models Qwen2.5-VL and InternVL3 have the best F1. These indicate that models with stronger foundational capabilities tend to perform better in our classification task.
3) Our framework consistently improves all baselines across accuracy and F1, demonstrating our method's effectiveness.  Notably, adding our method results in a more balanced prediction distribution, with SRatio values shifting closer to 50\%, indicating a more unbiased classification of strengtheners and weakeners. 


\textbf{Generation Task.}
Table \ref{table:gen_combined} presents the performance of VLMMs on the {generation task}. We observed that
1) Qwen2.5-VL and InternVL3 demonstrate the strongest generation ability for our generation task, as reflected in their high C1 and low C3/C4.
2) In contrast, VideoGPT+ and Video-ChatGPT perform poorly in reasoning precision, as evidenced by their higher C2, C3, and C4, suggesting that they frequently generate neutral updates or confuse Strengtheners with Weakeners. The potential reason may be their low instruction-following ability.
3) VideoGPT+ shows a higher C5 and C6, revealing that it tends to restate the premise or hypothesis instead of constructing meaningful updates. 
4) Our method significantly improves the proportion of correctly generated updates (C1) across all models, indicating the effectiveness of our generation method. 
5) Furthermore, the reduction in C3 suggests that our approach helps models better differentiate between strengthening and weakening updates.
6) Additionally, C8 sees a notable reduction across models, demonstrating improved coherence in generated updates. For example, VideoLLaMA\_V reduces nonsensical outputs from 34.41\% to 0.86\%. 
In addition, we also add \textbf{human evaluation} in the Appendix \ref{appendix:human_evaluation}.

\subsection{Ablation Study}

To better understand the contributions of different components in our classification framework, we conduct an ablation study by removing each module: (1) ASR-enhanced video descriptions, and (2) Counterfactual Thought reasoning. Table~\ref{table:cls_ablation} presents the performance impact of these ablations. Without ASR-enhanced descriptions, models tend to rely more on textual artifacts rather than comprehensively understanding the video content. For instance, VideoLLaMA2\_AV exhibits a recall drop from 72.21\% (w/ ours) to 44.83\% (w/o refined description), indicating that without enriched descriptions, models struggle to correctly classify entailment cases that require spoken content understanding.
Excluding Counterfactual Thought leads to a significant increase in errors where updates are misclassified as Strengtheners instead of Weakeners (or vice versa). This is reflected in the recall drop in VideoLLaMA2\_V from 77.89\% to 49.70\%, showing that the model without counterfactual reasoning guidance performs worse. 
These results confirm that our proposed enhancements effectively mitigate VLMMs’ limitations.

\begin{table}[t]
\centering
\small
\resizebox{\linewidth}{!}{%
\begin{tabular}{lcccc}
\toprule
\textbf{Model} & \textbf{Accuracy} & \textbf{Precision} & \textbf{Recall} & \textbf{F1} \\
\midrule
Video-ChatGPT + Ours & \textbf{77.79} & 84.77 & \textbf{67.75} & \textbf{75.31} \\
\quad w/o-Refined description & 74.75 & \textbf{86.97} & 58.22 & 69.74 \\
\quad w/o-Counterfactual & 71.91 & 84.18 & 53.96 & 65.76 \\
\midrule
VideoGPT+ + Ours & \textbf{77.16} & 75.87 & \textbf{79.72} & \textbf{77.74} \\
\quad w/o-Refined description & 77.06 & \textbf{79.87} & 72.41 & 75.96 \\
\quad w/o-Counterfactual & 72.99 & 75.51 & 68.15 & 71.64 \\
\midrule
VideoLLaMA2\_V + Ours & \textbf{77.89} & 82.51 & \textbf{70.79} & \textbf{76.20} \\
\quad w/o-Refined description & 76.37 & \textbf{85.52} & 63.49 & 72.88 \\
\quad w/o-Counterfactual & 71.60 & 88.45 & 49.70 & 63.64 \\
\midrule
VideoLLaMA2\_AV + Ours & \textbf{78.80} & 83.18 & \textbf{72.21} & \textbf{77.31} \\
\quad w/o-Refined description & 77.69 & 87.19 & 64.91 & 74.42 \\
\quad w/o-Counterfactual & 70.79 & \textbf{93.25} & 44.83 & 60.55 \\
\bottomrule
\end{tabular}
}
\caption{Ablation study on the classification task, evaluating the effects of ASR-enhanced descriptions, counterfactual reasoning, and rationale refinement. 
}
\label{table:cls_ablation}
\vspace{-3mm}
\end{table}

\section{Conclusion}

We introduce Defeasible Video Entailment, an extension of the video entailment task that requires dynamically revising inferences based on new updates. To enhance classification performance, we propose the Chain of Counterfactual Thought framework, which leverages counterfactual reasoning, ASR-enhanced Video Content Integration, and Rationale Refinement to reduce inference bias. For generation task, we have developed a framework that integrates an external ASR tool and LLM to produce coherent, contextually relevant updates aligned with the intended strengthener or weakener goal. 
Our approach significantly improves performance. 
Additionally, we introduce the \ours benchmark, providing a dataset with strengthener/weakener annotations to support future research. 
To address the limitations of traditional automatic metrics, we propose a novel LLM-based evaluation method for the generation task. 
Our work advances dynamic video reasoning and multimodal inference, laying the foundation for improving VLMMs’ video understanding capabilities.

\section{Limitations}

Despite our work’s significant contributions, there is still a limitation.
Our proposed models rely on large-scale black-box architectures, particularly LLMs, which, while effective, suffer from limited interpretability and hinder the ability to understand or trust the model’s internal reasoning process. 

\bibliography{acl_latex}
\clearpage
\setcounter{page}{1}
\appendix

\section{Human Evaluation} 
\label{appendix:human_evaluation}
To further validate the effectiveness of our method, we conducted a human evaluation, as previous research \citep{DBLP:journals/tois/JingSLZZN24} has shown that automatic evaluation may introduce unreliable errors in generation tasks. To optimize cost-efficiency, we randomly sampled 100 instances from two representative models, VideoGPT+ and VideoLLaMA2, including both their original and refined versions with our framework, resulting in a total of 400 samples. Each update was annotated by three evaluators, who classified it into one of the eight predefined categories. The final category assignment was determined by a majority vote among the annotators.    
The results are presented in Table~\ref{table:human_eval}.
From Table~\ref{table:human_eval}, we observe that our refinement method significantly increases the proportion of high-quality updates (C1), particularly for VideoGPT+, where the percentage of Good updates rises from 21\% to 88\%. Similarly, VideoLLaMA2 sees an improvement from 50\% to 84\%.
The decrease in C6 suggests that refined models generate fewer redundant updates and are better at introducing meaningful modifications.
These results confirm that our approach effectively enhances update generation by improving the logical consistency and informativeness of model outputs.
Furthermore, we conduct the \textbf{case study} for qualitative analysis in the Appendix \ref{sec:case_study}.

\begin{table}[h]
    \centering  
    \resizebox{0.45\textwidth}{!}{
    \begin{tabular}{lccccccccc}
        \toprule
        \textbf{Model} & 
        \textbf{w/Ours} &
        \textbf{C1} & 
        \textbf{C2} & 
        \textbf{C3} & 
        \textbf{C4} & 
        \textbf{C5} & 
        \textbf{C6} & 
        \textbf{C7} & 
        \textbf{C8} \\
        \midrule
        \multirow{2}{*}{VideoGPT+} &\xmark& 21 & 3 & 0 & 3 & 3 & 68 & 0 & \textbf{2} \\
         &\cmark & \textbf{88} & \textbf{1} & \textbf{1} & \textbf{0} & \textbf{4} & \textbf{2} & \textbf{2} & \textbf{2} \\
        \midrule
        \multirow{2}{*}{VideoLLaMA2} &\xmark  & 50 & 13 & 2 & 3 & 7 & 13 & 7 & 5 \\
        &\cmark & \textbf{84} & \textbf{2} & \textbf{0} & \textbf{1} & \textbf{5} & \textbf{1} & \textbf{5} & \textbf{2} \\
        \bottomrule
    \end{tabular}
    }
    \caption{Human evaluation results on update generation. Each percentage represents the proportion of updates falling into the respective category. For each model, its best results in each metric are highlighted in \textbf{bold}.}
    \label{table:human_eval}
\end{table}

\section{Prompts}
\label{sec:prompts}
In this section, we present all the prompts we used in this paper.
\subsection{Prompts for the Classification Task}
\label{subsec:prompt_cls_base}
We illustrate our prompt for the classification task for all the VLMMs in Figure \ref{fig:prompt_cls_base}.

\subsection{Prompt for baselines in the Generation Task}
\label{subsec:prompt_gen_base}
\begin{figure*}[htbp]
  \centering
  \begin{minipage}[t]{0.47\textwidth}
    \centering
    \includegraphics[width=\linewidth]{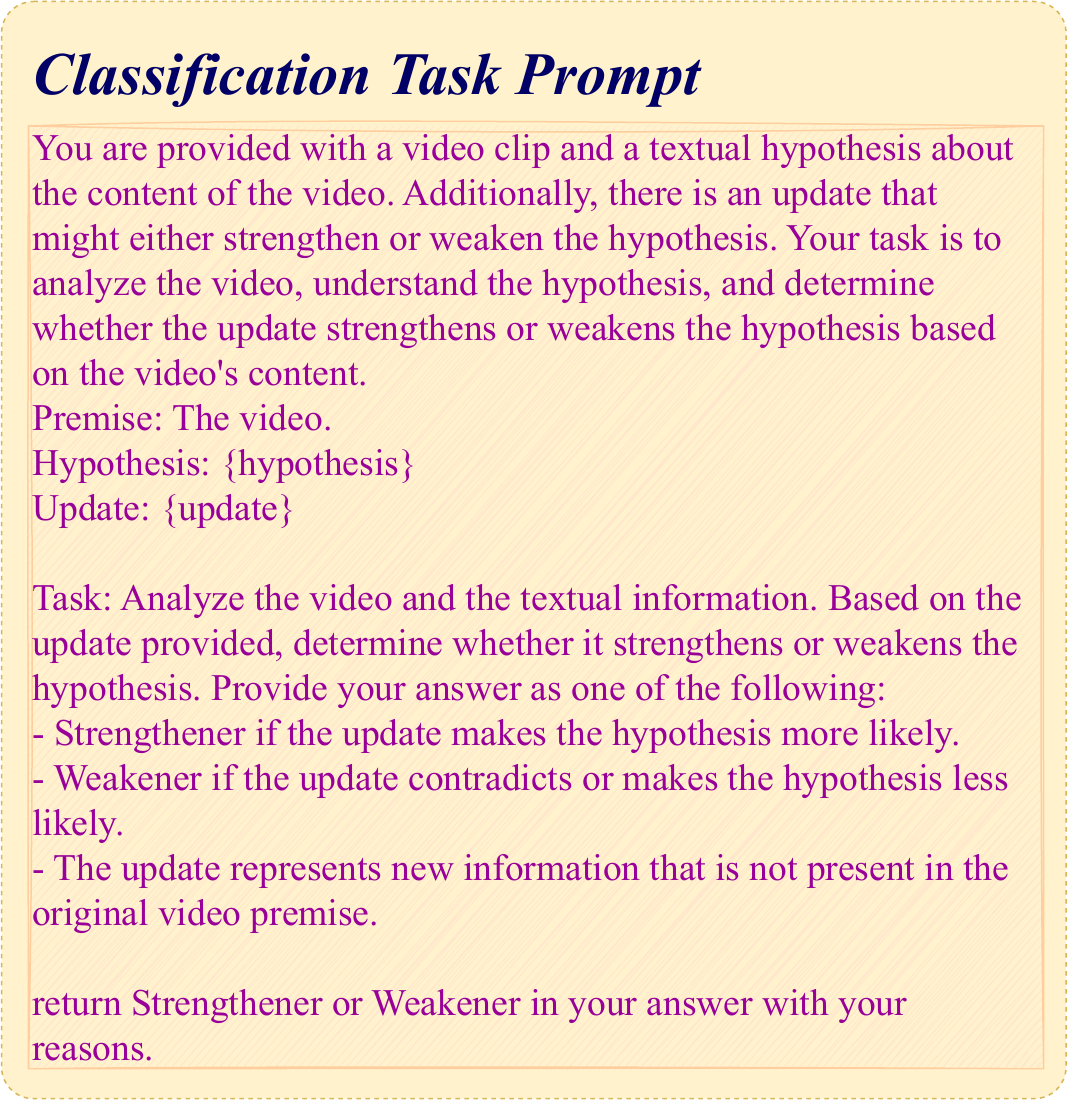}
    \caption{Prompt used for the Classification Task across VLMMs.}
    \label{fig:prompt_cls_base}
  \end{minipage}%
  \hfill
  \begin{minipage}[t]{0.47\textwidth}
    \centering
    \includegraphics[width=\linewidth]{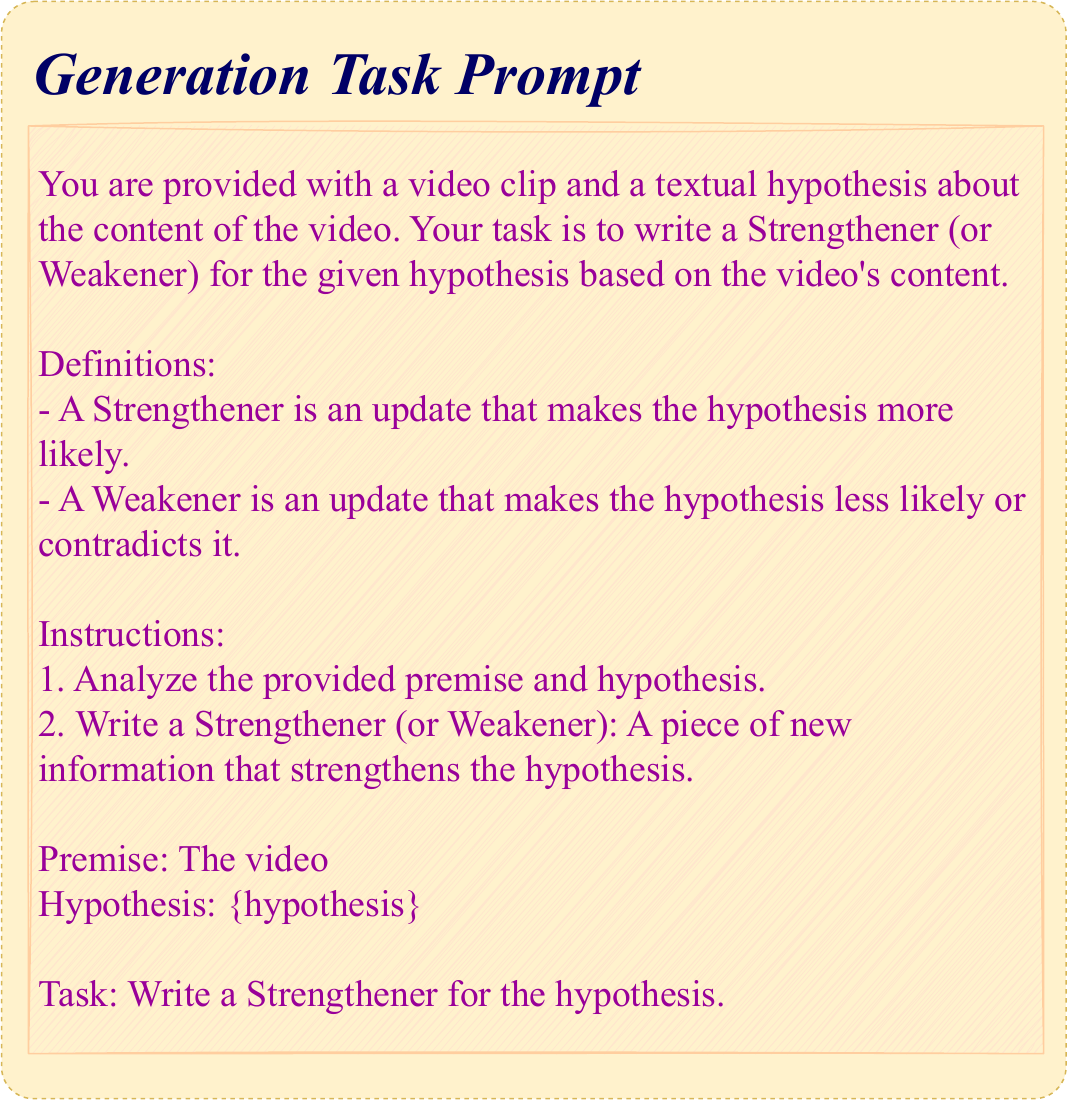}
    \caption{Prompt used for baselines in the Generation Task.}
    \label{fig:prompt_gen_base}
  \end{minipage}
\end{figure*}

Our prompt for the generation task, applied consistently across all models, is shown in Figure \ref{fig:prompt_gen_base}.

\subsection{Prompt for evaluating the generated updates}
\label{subsec:prompt_eval}
\begin{figure*}
    \centering
    \includegraphics[width=\linewidth]{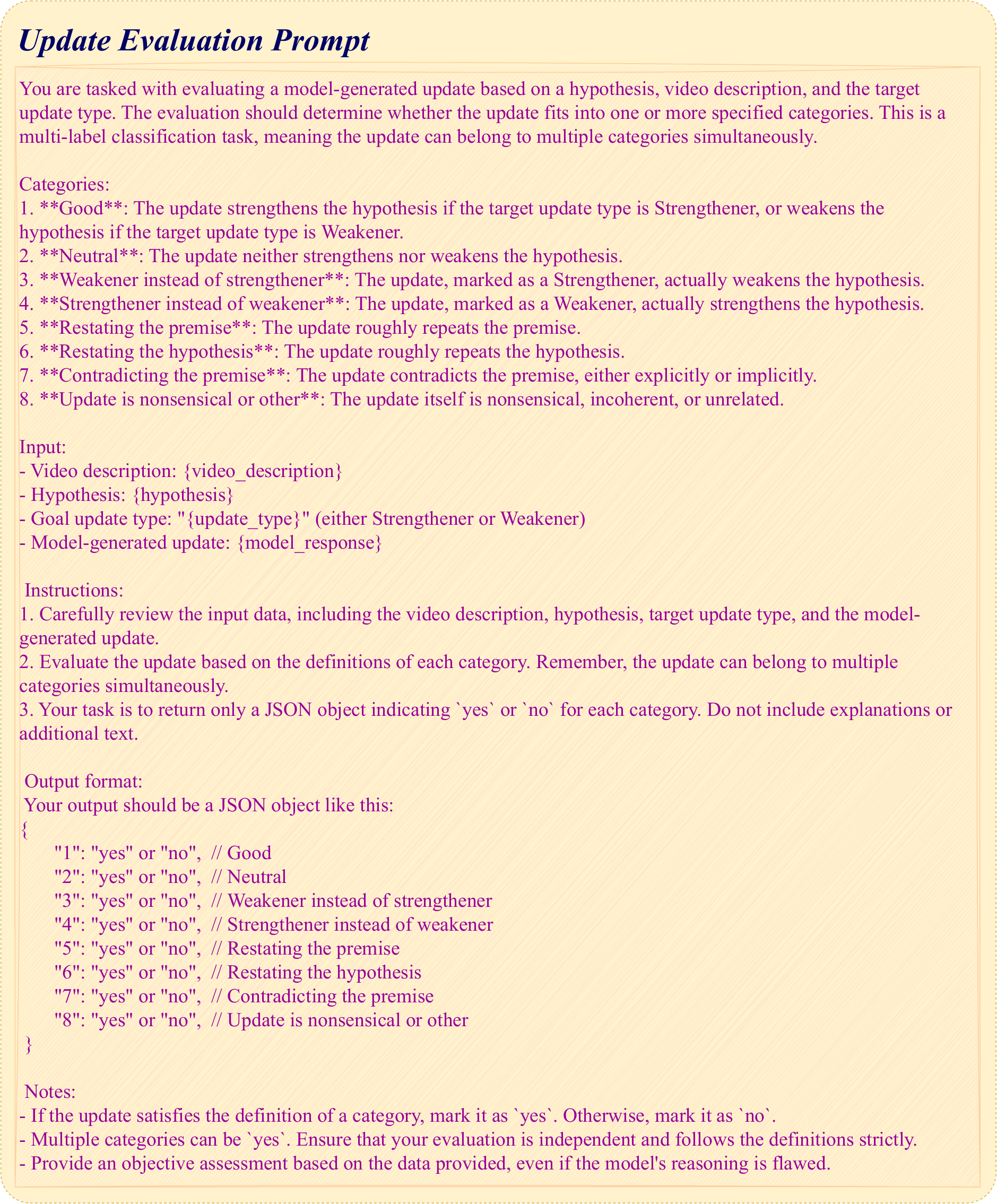}
    \caption{Prompt used for evaluating the generated updates.}
    \label{fig:prompt_eval}
\end{figure*}

Figure \ref{fig:prompt_eval} shows the prompt we used for evaluating the generated updates.

\subsection{Prompts in our Classification Framework}
\label{subsec:prompt_our_cls}
\begin{figure*}
    \centering
    \includegraphics[width=\linewidth]{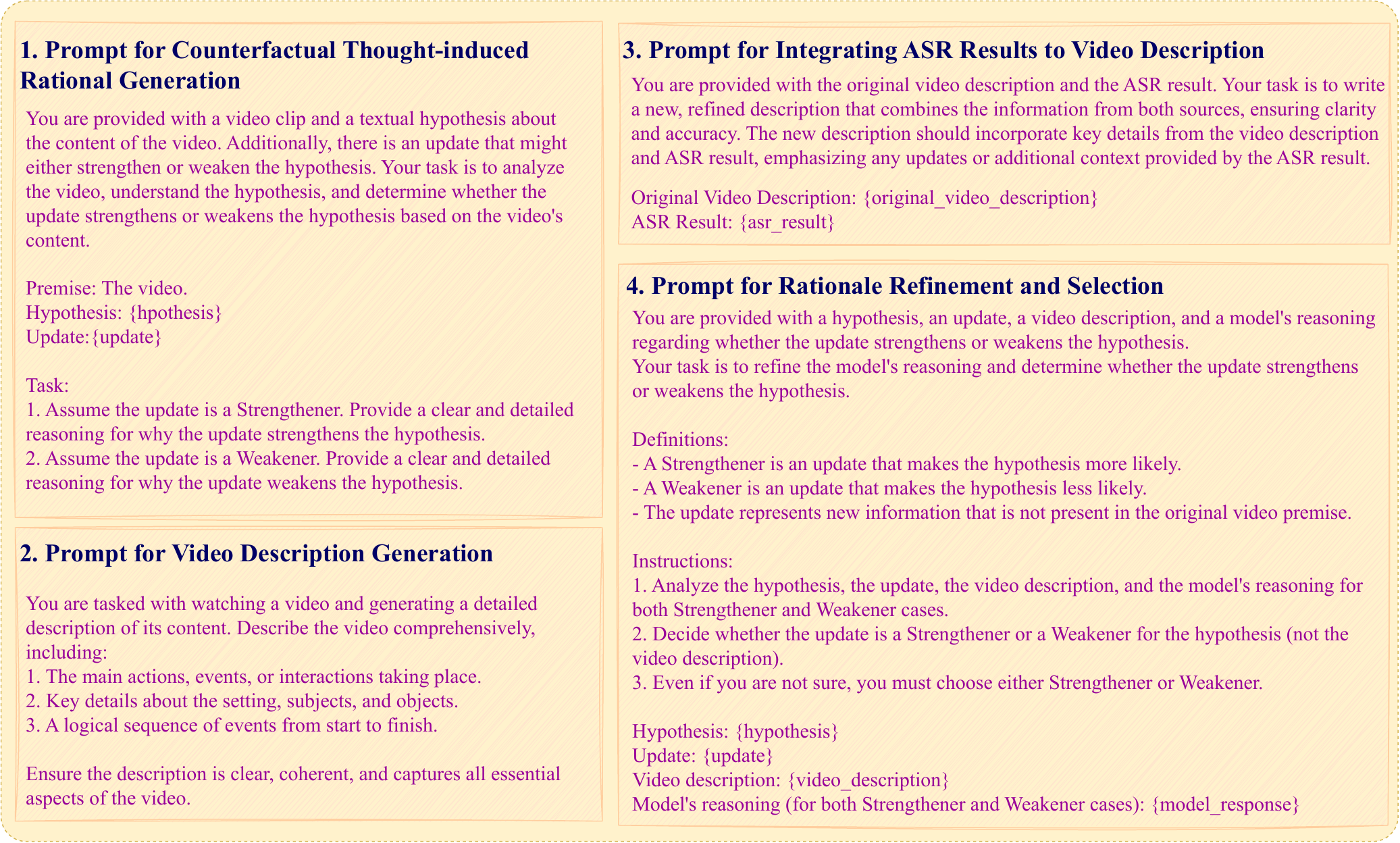}
    \caption{All prompts in our Classification Framework.}
    \label{fig:prompt_cls}
\end{figure*}

Figure \ref{fig:prompt_cls} illustrates all the prompts used in the Classification Framework. 

\subsection{Prompts in our Generation Framework}
\label{subsec:prompt_our_gen}
\begin{figure*}
    \centering
    \includegraphics[width=\linewidth]{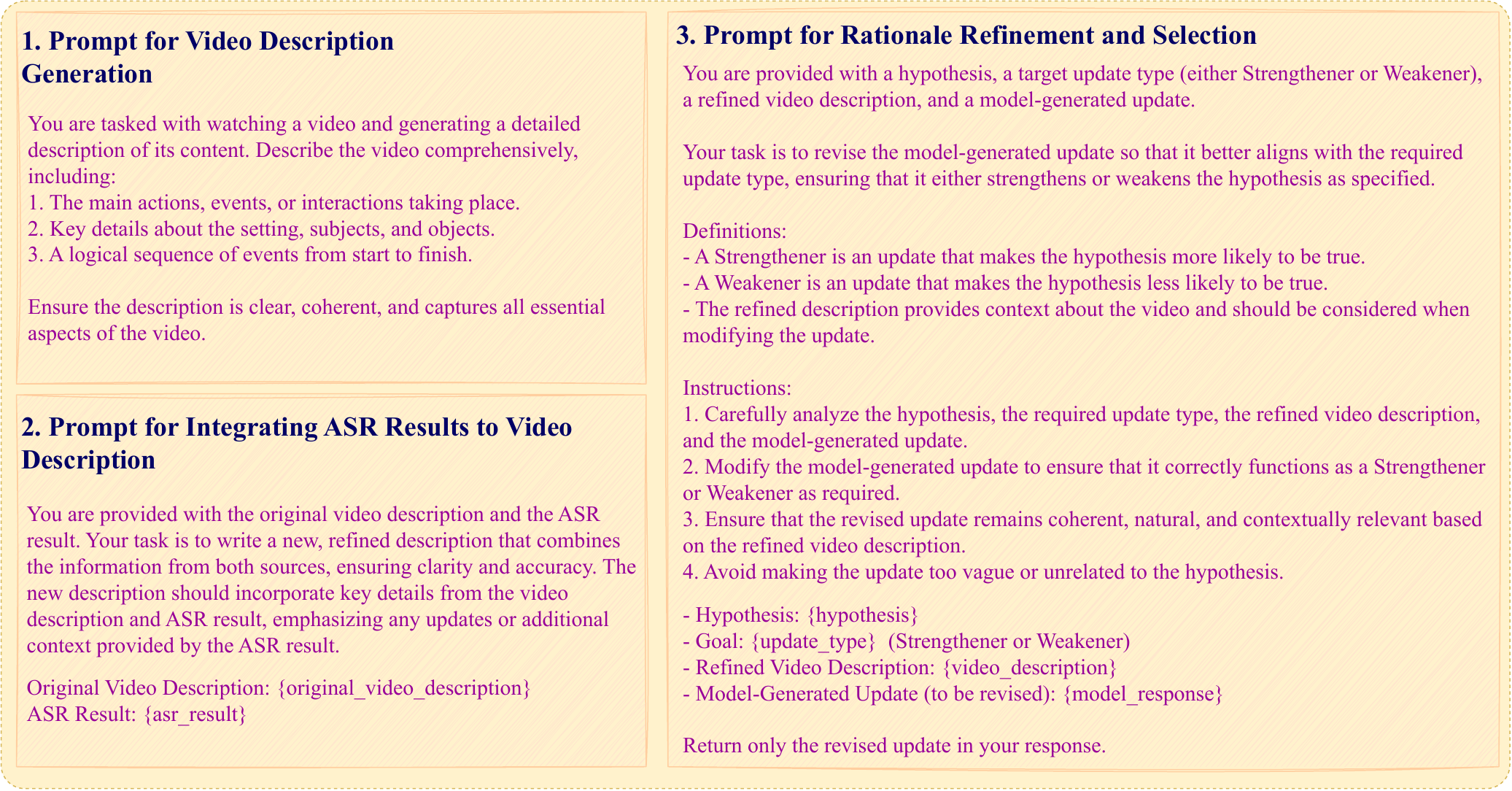}
    \caption{All prompts in our Generation Framework.}
    \label{fig:prompt_gen}
\end{figure*}

We provide the prompt used for the optimized method, as illustrated in Figure \ref{fig:prompt_gen}.

%

\section{Dataset Comparison}
\label{app:comparison}

Existing video understanding benchmarks like Charades \citep{DBLP:conf/eccv/SigurdssonVWFLG16}, InfiniBench \citep{DBLP:journals/corr/abs-2406-19875}, and Kinetics \citep{DBLP:journals/corr/KayCSZHVVGBNSZ17} provide extensive evaluations of model abilities to interpret and reason about fixed, predefined video scenes, such as recognizing actions, classifying events, and understanding temporal relationships. However, they do not assess a model’s ability to handle dynamic and nuanced semantic changes introduced by new, uncertain information
In contrast, DVidE explicitly evaluates a model’s ability to adapt its reasoning when presented with new evidence.

Natural Language Inference (NLI) datasets assess whether a hypothesis follows from a premise, typically classifying their relationship as entailment, contradiction, or neutral. Datasets such as SNLI \citep{DBLP:conf/emnlp/BowmanAPM15} and MNLI \citep{DBLP:conf/naacl/WilliamsNB18} focus solely on textual entailment without considering inference updates. To introduce defeasibility, datasets like $\delta$-NLI \citep{DBLP:conf/emnlp/RudingerSHBFBSC20} and $\delta$-CAUSAL \citep{DBLP:journals/corr/abs-2401-03183} provide updates that strengthen or weaken inference. However, these defeasible NLI datasets remain text-only and do not handle multi-modal reasoning. Furthermore, DVE~\citep{DBLP:conf/aaai/ZhangJG25} introduces image-based defeasible reasoning by replacing the textual premise with an image.
However, it remains limited to single images and lacks temporal reasoning. In contrast, DVidE explicitly introduces defeasibility in videos, allowing models to revise inferences as new temporal visual and auditory information flows in.

\section{Ethics Statement}
\label{app:ethics}

\subsection{Human Annotation Protocol}
\textbf{Recruitment \& Screening.}
We employed paid crowdworkers and graduate research assistants via an internal university pool. Candidates completed a qualification task with 20 multiple--choice items based on exemplar annotations; only those scoring \textbf{$\geq 85\%$} advanced to production.

\noindent\textbf{Compensation.}
Annotators were paid \$20/hour, above the local minimum wage and aligned with institutional guidance for human--subjects research compensation.

\noindent\textbf{Instructions \& Protections.}
Annotators received written guidelines describing the task (identify or compose \emph{strengthener}/\emph{weakener} updates for a given video--hypothesis pair), edge cases, and quality criteria. No personally identifying information (PII) was collected.

\subsection{Data Provenance, Licensing, and Intended Use}
Our benchmark augments \textbf{VIOLIN} video--hypothesis pairs with defeasible updates.
We do not collect new private media; instead, we reference timestamped clips retrievable from the original VIOLIN sources for transparency and traceability.
Released materials (textual updates/labels and retrieval metadata) are for \textbf{non--commercial research only}; users must comply with the terms and licenses of the underlying sources.

\subsection{Privacy, Safety, and Content Risk}
The dataset contains no PII (faces/voices occur within publicly released TV/movie material in VIOLIN).
Because scripted media may include harmful or offensive themes, we provide a \emph{content disclaimer} and advise against use in safety--critical or sensitive settings (e.g., hiring, policing, healthcare).
Downstream users should conduct separate safety and bias assessments before any deployment.

\subsection{Release Protocol}
We will release our data after this work has been accepted, under a \textbf{Creative Commons Attribution–NonCommercial 4.0 International (CC BY-NC 4.0)} license. 
This license permits redistribution and adaptation for non-commercial research purposes with appropriate credit.

\section{Case Study}
\label{sec:case_study}
\begin{figure*}[t]
    \centering
    \includegraphics[width=0.90\linewidth]{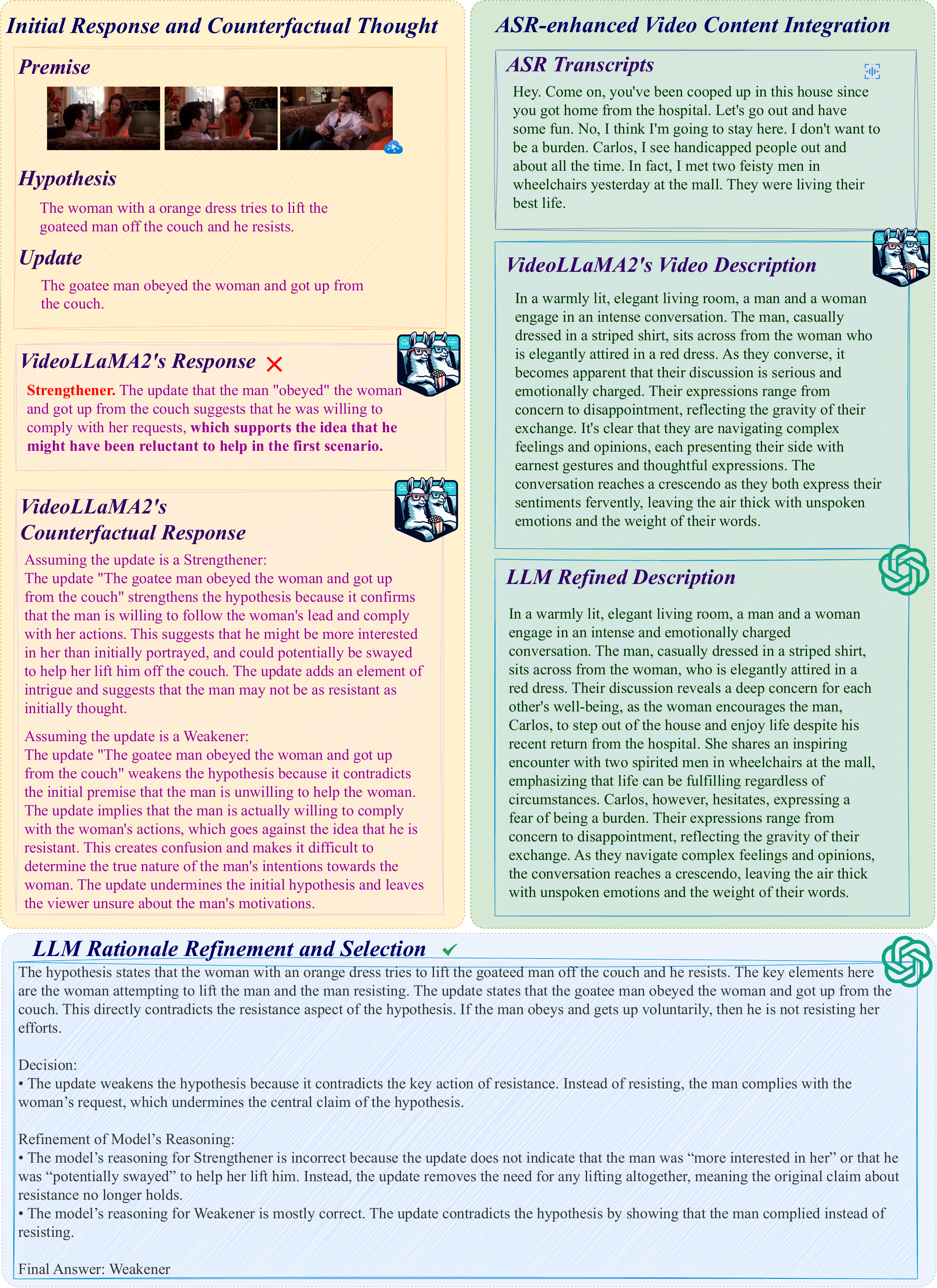}
    \caption{Enhancing VideoLLaMA2 in \ours: A Classification Task Example}
    \label{fig:case1}
\end{figure*}

While the quantitative results presented demonstrate overall performance trends, they do not fully capture how different models behave in  \ours.
To learn the qualitative performance of the VLMMs, we show representative cases in Figure \ref{fig:case1}. 
In the first case, the hypothesis explicitly states that the man ``resisted'', while the update describes a situation where he ``obeyed'' the woman’s request. Since resistance and obedience are conceptually opposed, the update should be categorized as a Weakener. However, VideoLLaMA2 classifies it as a Strengthener, suggesting that the model has difficulty understanding the logical relationship between the update and the hypothesis. With the counterfactual thought, the model successfully realized that the relationship between ``resisted'' and ``obeyed''. With the ASR transcripts, the refined description incorporates crucial contextual details that aid in a more accurate interpretation of the scene. The ASR output reveals that the man expresses hesitation about going outside, citing his recent hospitalization and concerns about being a burden. The LLM rationale refinement and selection corrects the initial misclassification by eliminating speculative reasoning and focusing on direct textual contradiction. Initially, VideoLLaMA2 incorrectly categorized the update as a Strengthener by assuming that obedience might indicate increased interest or willingness to help. However, the refined rationale correctly identifies that the update negates the key element of “resistance” in the hypothesis—if the man obeys and gets up voluntarily, then there is no longer any resistance, contradicting the original claim. This correction ensures that classification is based strictly on entailment logic rather than inferred motivation, demonstrating how rationale refinement improves logical consistency in multi-turn reasoning.

\begin{figure*}[t]
    \centering
    \includegraphics[width=\linewidth]{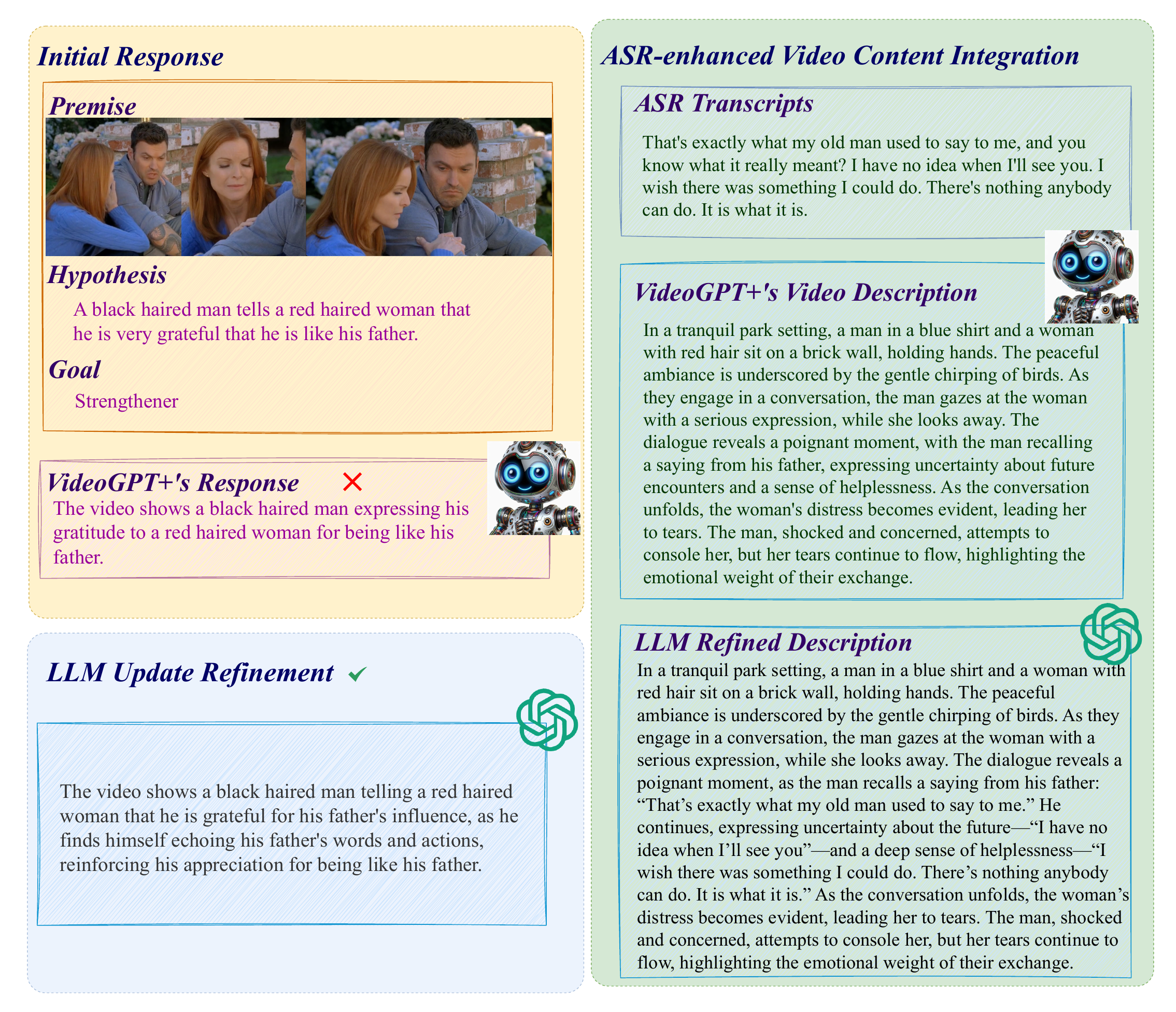}
    \caption{Enhancing VideoGPT+ in \ours: A Generation Task Example}
    \label{fig:case2}
\end{figure*}

A similar case is illustrated in Figure \ref{fig:case2}. Here, the model attempts to generate a strengthener, but VideoGPT+‘s response merely paraphrases the hypothesis rather than providing additional supporting information. By applying our method, the update has been successfully refined to incorporate new, relevant details that are absent from the original video premise while remaining closely aligned with its content—specifically, by echoing the father’s words and actions.

\section{Usage of LLMs}
We made limited use of large language models—specifically ChatGPT and GitHub Copilot—for
language polishing and routine coding assistance (autocompletion, refactoring, and boilerplate generation). LLMs were not used to generate research ideas, experimental results, or claims. No non-public or sensitive data was shared with these tools. All suggested text and code were independently reviewed, tested, and edited by the authors, who remain responsible for the content.

\end{document}